\documentclass{INTERSPEECH2023}

\interspeechcameraready 

\usepackage{array}
\usepackage{graphicx}
\usepackage{amsmath}
\usepackage[capitalize,noabbrev]{cleveref}
\usepackage{subcaption}
\usepackage{booktabs}
\usepackage{float}
\usepackage{multirow, tabularx}
\usepackage{nth}
\usepackage{url}

\title{Adversarial learning of neural user simulators for dialogue policy optimisation}

\name{Simon Keizer$^1$, Caroline Dockes$^2$ \thanks{A preliminary version of this research was carried out by Caroline Dockes as part of her MPhil in Machine Learning and Machine Intelligence \cite{Dockes2021UsimGAN}, hosted by Toshiba Europe Limited.}, Norbert Braunschweiler$^1$, Svetlana Stoyanchev$^1$, Rama Doddipatla$^1$}
\address{
  $^1$Toshiba Europe Limited, Cambridge Research Lab, UK \\
  $^2$University of Cambridge, Department of Engineering, UK
}
\email{firstname.lastname@toshiba.eu}

\begin{document}

\maketitle

\begin{abstract}
Reinforcement learning based dialogue policies are typically trained in interaction with a user simulator.  To obtain an effective and robust policy, this simulator should generate user behaviour that is both realistic and varied.  Current data-driven simulators are trained to accurately model the user behaviour in a dialogue corpus.  We propose an alternative method using adversarial learning, with the aim to simulate realistic user behaviour with more variation.  We train and evaluate several simulators on a corpus of restaurant search dialogues, and then use them to train dialogue system policies.  In policy cross-evaluation experiments we demonstrate that an adversarially trained simulator produces policies with 8.3\% higher success rate than those trained with a maximum likelihood simulator.  Subjective results from a crowd-sourced dialogue system user evaluation confirm the effectiveness of adversarially training user simulators.
\end{abstract}


\section{Introduction}

User simulators have been used successfully as a technique for training, testing and evaluating spoken dialogue systems.  The agenda-based user simulator \cite{schatzmann-etal-2007-agenda} in particular is widely used for training dialogue policies using reinforcement learning.  However, dialogue systems that have been optimised in this way do not perform as well in interaction with real users \cite{Schatzmann-ea_ASRU-2005, kreyssig-etal-2018-neural}.  Therefore, in order to generate more realistic user behaviour, statistical models that can be trained from data have been developed.  In recent years, neural user simulators have emerged with promising results \cite{asri16_interspeech, crook17_interspeech, kreyssig-etal-2018-neural, gur18-slt-usim, lin-etal-2021-domain, Lin_etal_2022_sigdial}.

However, data-driven user simulators rely on the range and variation of user behaviour patterns covered in the training data.  As a consequence, policies trained with such simulators might not be sufficiently prepared for new users, especially if the training corpus is small.  To some extent, regularisation can be applied to better generalise beyond the training data, but this is typically controlled by how well the model matches the development data, which often is very similar in nature.  To further analyse this, we look at different training methods and how they impact both performance on corpus data (i.e., through intrinsic, or direct evaluation) and performance of the trained policies (i.e., through extrinsic, or indirect evaluation).

Since a user simulator is essentially a generative model that is required to produce realistic and diverse user responses, a Generative Adversarial Network (GAN) \cite{goodfellow2014generative} could be a promising architecture to consider for this.  GANs and other adversarial learning methods have been used successfully in image generation \cite{Shamsolmoali-ImgSynthAdv-2021, Frolov-etal-AdvTxt2Img-2021} and have also found their way into the fields of natural language processing and dialogue modelling \cite{li-etal-2017-adversarial, liu-lane-2018-adversarial}, but have not yet been applied to conversational user simulation.

In the next section, we describe our semantic level neural user simulation model and how it is extended into a GAN by introducing a discriminator model that is trained simultaneously with the main simulator.  Then, we describe our experiments with different training methods, distinguishing between standard Maximum Likelihood Estimation (MLE) and GAN-based training, and varying hyper-parameter settings and levels of pre-training.  Next, we explain how the trained simulation models are used to train dialogue policies and then cross-evaluated.  Finally, we present results of a crowd-sourced user evaluation, comparing three selected sets of policies, corresponding to the best MLE simulator, the best GAN simulator, and our agenda-based simulator these policies were trained with.  After discussing related work on neural user simulation and adversarial learning in more detail, the paper is wrapped up with conclusions and directions for future work.

\begin{figure*}[htb]
    \centering
    \begin{subfigure}[b]{.8\textwidth}
    \centering
    \includegraphics[width=.95\textwidth]{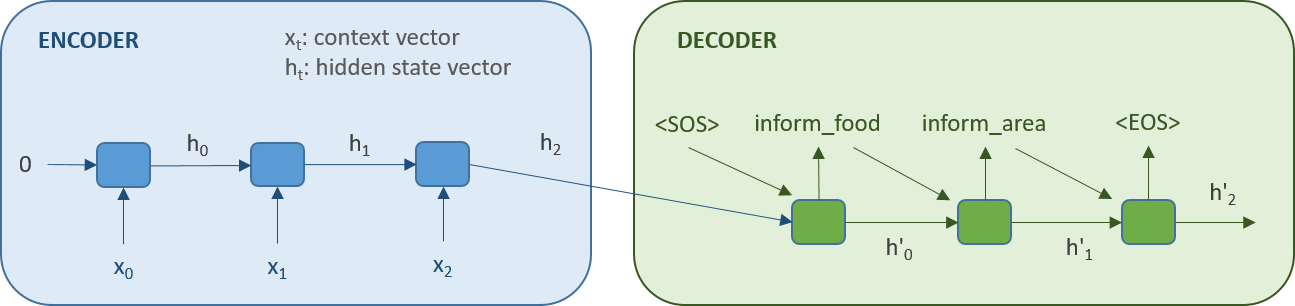}
    \end{subfigure}
    \\[2ex]
    \begin{subfigure}[b]{.8\textwidth}
    \centering
    \includegraphics[width=.95\textwidth]{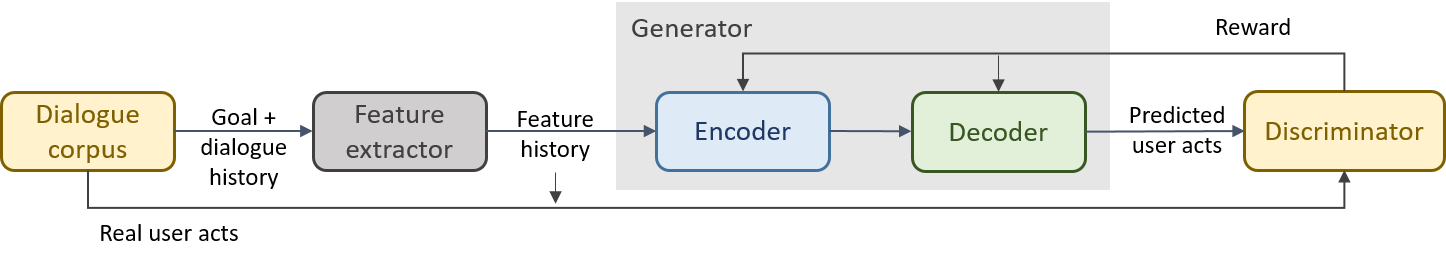}
    \end{subfigure}
    %
    \begin{subfigure}[b]{.8\textwidth}
    \centering
    \includegraphics[width=.95\textwidth]{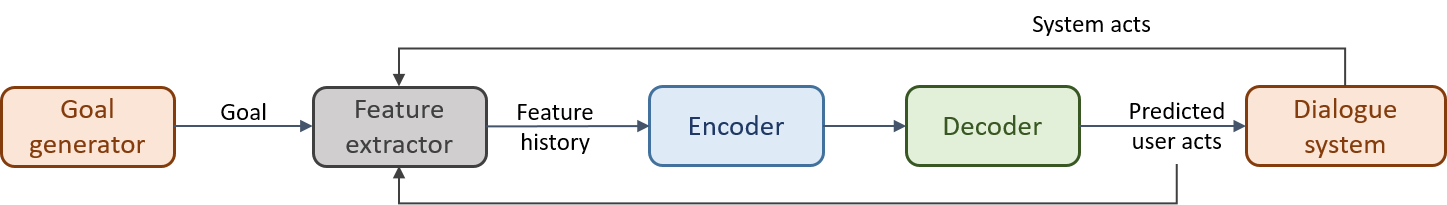}
    \end{subfigure}
    %
    \caption{Overview of the neural user simulator: seq2seq model (top), GAN training (middle), deployed simulator (bottom).}
    \label{fig:gan-usim}
\end{figure*}

\section{Seq2seq neural simulator model}\label{sec:nus}

The neural user simulator is implemented as a sequence-to-sequence (seq2seq) model, consisting of an LSTM encoder and an LSTM decoder (both with a single hidden layer of size 32); see \cref{fig:gan-usim}.  At every turn, the encoder provides a hidden representation $h_t$ of the dialogue so far, based on the hidden representation of the previous turn $h_{t-1}$ and the dialogue context vector $x_t$ for the current turn.  Following \cite{asri16_interspeech}, the context vector is provided by a Feature Extractor and encodes the last user act, the last system act, and any inconsistencies between the system act and the user goal (e.g., when the user goal is to find a cheap restaurant and the system offers an expensive one).  The user goal is taken from the data (during training) or generated by a goal generator (when running the simulator against a dialogue system).  The decoder takes the latest hidden vector as input and generates a sequence of user acts (starting with the start-of-sentence symbol \textlangle{}SOS\textrangle{}) that together form the user response for the current turn.  Generation stops when the end-of-sentence symbol \textlangle{}EOS\textrangle{} is generated, or when three user acts have been generated.

\subsection{Adversarial learning method}\label{sec:nus-gan}

The neural architecture described above can be extended to a Generative Adversarial Network (GAN) by viewing the seq2seq model as a Generator and introducing a new model, the Discriminator, which classifies the user responses from the Generator as being simulated or real.  Generator and Discriminator are trained simultaneously, where the Discriminator is trained on both real user responses from the training corpus and simulated responses from the Generator, and the Generator uses the classification probabilities (of the user response being real) from the Discriminator as reward signal for training.  Hence, an alternative method for training the seq2seq Generator is introduced, which uses reinforcement learning in interaction with the Discriminator.

The Discriminator is currently modelled as a 2-layer feedforward neural network (with a hidden layer of size 64) that takes as input the latest context vector for the current dialogue and the user act sequence (either from the real user data or simulated by the Generator), and has a softmax layer (of size 2) at the output to produce probabilities for the user response being simulated or real.  The Generator is trained using the REINFORCE algorithm \cite{williams1992simple}, following previous work on adversarial learning in discrete application domains \cite{demasson2019}.  We also apply `teacher forcing' \cite{williams1989teacher} in 50\% of user turns, i.e., when generating a user response we sometimes feed the real user acts from the corpus used in the same context as input to the next generation step, rather than the predicted act.

An overview of the overall architecture is shown in \cref{fig:gan-usim}.  When training and evaluating on a dialogue corpus (middle figure), the output of the decoder is passed to the Discriminator, which then passes its output as a reward back to the Generator.  The Generator uses the received rewards to update the parameters of both decoder and encoder models.  The dialogue contexts that feed into the encoder are taken from the corpus; the real user responses that feed into the Discriminator are also taken from the corpus.  When deploying the user simulator (bottom figure), the output of the decoder is passed to the dialogue system, which generates the next system dialogue act, to be fed back into the encoder (via the Feature Extractor).  In this case, the user goals are generated randomly from the domain ontology by a Goal Generator.

\begin{table*}[htb]
\centering
\begin{tabular}{lcccccccc}
\toprule
       & \multicolumn{3}{c}{\bf Generator} & \multicolumn{2}{c}{\bf Discriminator} & \multicolumn{3}{c}{\bf Test set evaluation} \\
{\bf Simulator} & {\it pre-training} & $\alpha$ & $\lambda$ & $\alpha$ & $\lambda$ & {\it F-score} $\uparrow$ & {\it KL-div} $\downarrow$ & {\it Entropy} $\uparrow$ \\
\midrule
MLE-tuned       &   --       &  1e-04 & 1e-03  & \multicolumn{2}{c}{--}  &  49.46  &  \textbf{0.353}  &  1.547 \\
MLE-default     &   --       &  1e-03 & 0      & \multicolumn{2}{c}{--}  &  \textbf{49.67}  &  0.466  &  1.484 \\
GAN-tuned       &   0 epochs &  1e-04 & 1e-03  &    5e-04  &  1e-05      &  33.55  &  0.874  &  2.176 \\
GAN-pre1-tuned  &   1 epochs &  1e-04 & 1e-03  &    5e-04  &  1e-05      &  34.01  &  0.890  &  \textbf{2.207} \\
GAN-pre10-tuned &  10 epochs &  1e-04 & 1e-03  &    5e-04  &  1e-05      &  49.32  &  0.567  &  1.835 \\
GAN-pre30-tuned &  30 epochs &  1e-04 & 1e-03  &    5e-04  &  1e-05      &  48.52  &  0.517  &  1.871 \\
\bottomrule
\end{tabular}
\caption{Overview of user simulation models selected for policy optimisation.}
\label{tab:usim-vars}
\end{table*}

\begin{table*}[htb]
  \centering
  \begin{tabular}{l| ccc| ccc| ccc}
    \toprule
                    & \multicolumn{3}{c|}{\bf Against Neural Simulators} & \multicolumn{3}{c|}{\bf Against ABUS} & \multicolumn{3}{c}{\bf Overall} \\
    {\bf Policy}    & {\it Succ} $\uparrow$ & {\it Rew} $\uparrow$ &  {\it Rank} $\downarrow$ &  {\it Succ} $\uparrow$ & {\it Rew} $\uparrow$  & {\it Rank} $\downarrow$ & {\it Succ} $\uparrow$ & {\it Rew} $\uparrow$ & {\it Rank} $\downarrow$ \\
    \midrule
    MLE-tuned       & 76.41 & 65.83 & 4 & 72.74 & 62.98 & 4 & 74.57 & 64.41 & 4 \\
    MLE-default     & 76.06 & 64.86 & 5 & 63.18 & 53.45 & 5 & 69.62 & 59.15 & 5 \\
    GAN-tuned       & {\bf 77.43} & 66.52 & 2 & 74.76 & 63.77 & 3 & 76.10 & 65.14 & 3 \\
    GAN-pre1-tuned  & 77.15 & {\bf 67.17} & 1 & {\bf 88.68} & {\bf 78.27} & 2 & {\bf 82.91} & {\bf 72.72} & 2 \\
    GAN-pre10-tuned & 74.50 & 63.40 & 7 & 56.12 & 45.28 & 6 & 65.31 & 54.34 & 6 \\
    GAN-pre30-tuned & 75.74 & 63.97 & 6 & 52.86 & 41.65 & 7 & 64.30 & 52.81 & 7 \\
    \midrule
    ABUS            & 75.06 & 66.07 & 3 & {\bf 96.70} & {\bf 88.24} & 1 & {\bf 85.88} & {\bf 77.15} & 1 \\
    \bottomrule
  \end{tabular}
  \caption{Policy cross-evaluation results summary in terms of average success rates ({\it Succ}), average dialogue length ({\it Len}) and average reward ({\it Rew}).  The ranking orders ({\it Rank}) are based on the average rewards.}
  \label{tab:pol-xeval-summ}
\end{table*}

\section{User simulator training}\label{sec:nus_tra}

\subsection{Training setup}

For training and evaluating simulation models, we use the DSTC-2 corpus \cite{henderson-etal-2014-second} of transcribed and annotated spoken dialogues in the restaurant search domain.  For all neural network models we use the Adam optimiser \cite{kingma2015adam} to train the parameters.  Besides distinguishing between MLE training and GAN training, we also look at the effects of tuning the learning rate ($\alpha$) and weight decay ($\lambda$) hyper-parameters.  The trained models are evaluated in terms of F-score for the predicted user act sequences, KL-divergence between the real and simulated user response distributions (conditional on the context and then averaged over all contexts), and average entropy of the distributions from which the model samples when generating user response acts.  The latter is a useful measure of variation in user behaviour, which is important for training dialogue policies.  It is computed by taking the entropy of the model's user act distribution used for each generation step in each turn, and then averaged across the data (the test set in this case).


For conventional Maximum Likelihood Estimation (MLE) training, we minimise the negative log-likelihood of the training data given the model parameters.  We tune the learning rate and weight decay hyper-parameters using the development set, and then evaluate two models on the test set: MLE-tuned (trained with tuned hyper-parameters) and MLE-default (trained with default hyper-parameters).  

For training simulation models using the proposed GAN architecture, we experiment with different levels of MLE pre-training and hyper-parameter settings for both Generator and Discriminator training.  The Generator is pre-trained for up to 30 epochs, and the Discriminator is then pre-trained on the same number of epochs, taking user responses from that Generator as well as from the corpus as input.  During adversarial training, Generator and Discriminator are trained simultaneously.


\subsection{Direct evaluation results}

Since existing corpus-based metrics are too limited to fully predict how well a dialogue policy trained with a given simulator would perform, several of the trained simulators discussed so far are selected for the dialogue policy optimisation and (cross-) evaluation experiments, including 2 MLE based and 4 GAN-based models, as well as an agenda-based simulator.  An overview of the neural models is given in \cref{tab:usim-vars}, in terms of hyper-parameter settings for the Generator and Discriminator models (learning rate $\alpha$ and weight decay $\lambda$) and the number of pre-training epochs, along with the test set evaluation results.  As expected, the MLE-based simulators have the best F-score and KL-divergences, but their entropy is lower than those of the GAN-based models.  What can also be noticed is that the GAN-models that use more pre-training get better F-scores and KL-divergences, but lower entropies.  We will see in the policy cross-evaluation experiments that such observations have predictive value, but the link between direct and indirect evaluations is not a straightforward one.

\section{Policy optimisation and cross-evaluation}\label{sec:pol-tra-xeval}

In this experiment, we train policies with each of the selected user simulators, and then evaluate the resulting policies against all simulators, thus obtaining a matrix with results for every combination of training and evaluation simulator.  In the results \cref{tab:pol-xeval-summ}, the rows correspond to the simulators used for training, and the columns are grouped according to the type of simulator used for evaluation (neural, agenda-based, and overall).

For optimising the policies, Monte Carlo Control reinforcement learning with linear value function approximation is used \cite{Sutton1998}.  For each user simulator, 5 training runs over 40k dialogues at 25\% semantic error rate
are carried out, using linearly decaying learning rate and Boltzmann exploration temperature.  The reward function consists of +100 for a successful dialogue, -1 per turn, and -5 for violating a social convention (e.g., not responding to a goodbye act), provided by the simulator at each turn.  The policies are trained to maximize the long-term cumulative reward (across turns).  All final policies are evaluated over 1000 dialogues at 25\% semantic error rate against all simulators.  In each setting, we report the success rate
and average reward, averaged across the 5 trained policies.  The resulting scores are then averaged across all neural user simulators used in the evaluations, which finally are averaged with the results from evaluating against the agenda-based simulator.

The results in \cref{tab:pol-xeval-summ} show that out of the neural simulators, the best results are achieved with the policies from the GAN-pre1-tuned simulator, both against the neural simulators and against the ABUS simulator.  Overall, the best GAN-based policies outperform the best MLE-based policies by 8.3\% success rate.  The policies that were trained with the ABUS simulator achieve the best results overall, but this is mainly due to their strong performance against the ABUS simulator itself, i.e., the same simulator used for training.  It also seems that using more pre-training results in policies that perform worse, especially against the ABUS simulator.  This may be due to them being exposed to user behaviour with lower variation during training (lower entropy, as shown in \cref{tab:usim-vars}).  Due to the largely non-stochastic nature of the ABUS simulator, the differences in cross-evaluation performance may be exaggerated, as this simulator can get stuck in an infinite loop during a conversation, where simulators that have more variation in their behaviour are able to escape such situations.


\section{Human user evaluation}\label{sec:usr-eval}

To assess how realistic and varied the simulated user behaviour is and, therefore, how predictive the cross-evaluation results are, a crowd-sourced user evaluation was carried out.  We compared three dialogue system variants, defined by which user simulator was used to train their dialogue policies: one variant based on the best MLE-based simulator (MLE-tuned), another based on the best GAN-based simulator (GAN-pre1-tuned), and the third one based on the ABUS simulator.  To obtain representative results, all 5 policies that were trained in each of these three conditions were evaluated, where for each new dialogue one of the policies was selected, based on a round-robin system.

In the experiments, subjects were given a scenario to follow during their conversation with the system (e.g., ``You want to find a restaurant in the moderate price range that serves Indian food, and get the address.'').  After interacting with the system, they were asked to fill out a short questionnaire, in which the subject was required to state their opinion on 6 statements about the conversation, in the form of either a binary Yes or No (Q1 and Q2), or on a 6 point Likert scale (Q3 to Q6), ranging from `Strongly disagree' to `Strongly agree':

\smallskip

\begin{itemize}
\setlength\itemsep{0.5em}
\item[] Q1: The system recommended a restaurant that matched my constraints. [Y/N]
\item[] Q2: I got all the information I was looking for. [Y/N]
\item[] Q3: The system understood what I was saying. [1-6]
\item[] Q4: The system recognised my speech well. [1-6]
\item[] Q5: The system's responses were appropriate. [1-6]
\item[] Q6: The conversation felt natural. [1-6]
\end{itemize}

\medskip

The subjects used a web-based GUI that employs the Google Web Speech API for speech recognition and synthesis, and a server for sending recognised user utterances and receiving system responses from our dialogue system.  The dialogue system itself has components for language understanding and state tracking \cite{stoyanchev-etal-icassp2021}, action selection \cite{keizer-etal-asru2021}, and template-based natural language generation.

\begin{table}[h]
  \centering
  \begin{tabular}{lcccccccc}
  \toprule
  {\bf Policy} & {\bf Num} & {\bf Average} & {\bf Q1 [\%]} & {\bf Q2 [\%]} \\
  {\bf Usim} & {\bf Dials} & {\bf Length} & {\bf Found Venue} & {\bf DialSuccess} \\
  \midrule
  MLE   & 209 &      5.94  (3.92) &      87.56  (2.29) &      86.60  (2.36) \\
  GAN   & 204 & {\bf 5.36} (3.13) & {\bf 89.22} (2.18) &      88.24  (2.26) \\
  ABUS  & 200 &      5.49  (3.59) &      89.00  (2.22) & {\bf 88.50} (2.26) \\
  \bottomrule
  \end{tabular}
  
  \bigskip
  
  \begin{tabular}{lcccccccc}
  \toprule
  {\bf Policy} &    {\bf Q3 [1-6]} &   {\bf Q4 [1-6]}  &   {\bf Q5 [1-6]}  &   {\bf Q6 [1-6]}  \\
  {\bf Usim}   &    {\bf Underst}  &   {\bf Recogn}    &   {\bf SysResp}   &   {\bf Natural}   \\
  \midrule
  MLE          &      4.61  (1.58) &      4.85  (1.45) &      4.78  (1.43) &      4.44  (1.59) \\
  GAN          & {\bf 4.93} (1.34) & {\bf 4.91} (1.29) & {\bf 5.04} (1.30) & {\bf 4.70} (1.41) \\
  ABUS         &      4.79  (1.40) &      4.84  (1.38) &      4.96  (1.37) &      4.58  (1.50) \\
  \bottomrule
  \end{tabular}
\vspace{2mm}
  \caption{User evaluation results: average length in terms of system turns per dialogue; Q1-6 average scores from the questionnaire (standard deviations in brackets).}
  \label{tab:usr-eval}
\end{table}

\vspace{-5mm}
\subsection{User evaluation results}

The results from the user evaluation are summarised in \cref{tab:usr-eval}.  Overall, the scores are quite similar between the three conditions, though in most cases, the GAN based policies get the best score and the MLE based policies the worst.  The policy cross-evaluation results against the neural simulators appear to be more predictive than those based on the agenda based simulator, at least in terms of relative performance.  The agenda-based simulator evaluations suggested big differences in performance between the policies, but we did not observe this in the human evaluation.  The perceived success rates from Q1 and Q2 turn out to be much higher than the objective success rates obtained in simulation (over 10\% difference), suggesting that the MTurk workers were more effective than simulated users in completing their tasks, or that they may have overrated the systems in questions Q1 and Q2. Further data analysis would have to confirm this.

\section{Related work}\label{sec:rel-work}

Although agenda-based simulators \cite{schatzmann-etal-2007-agenda} have been very effective for training dialogue policies in limited domains, they are rule-based models that do not scale well to larger, more complex domains.  Furthermore, previous research suggests that data-driven simulators can be used to train policies that perform better in interactions with real users \cite{schatzmann2009hiddenagenda,kreyssig-etal-2018-neural}.

Our baseline seq2seq model, trained using maximum likelihood, is similar to the one proposed in \cite{asri16_interspeech}, which was also trained on the DSTC-2 corpus, but only reported direct evaluation results in terms of F-score.  A seq2seq model that operated on the word level (text input and output) was proposed in \cite{crook17_interspeech}.  Evaluation results were presented in terms of model perplexity and human judgements of generated dialogues.  A seq2seq model for word-level user simulation that was used for training dialogue policies was presented in \cite{kreyssig-etal-2018-neural}, which included a policy cross-evaluation and a real user evaluation, but no direct evaluation results on the corpus, and only one neural simulator model was selected for policy optimisation.  In \cite{tseng-etal-2021-transferable}, a combination of supervised pre-training, fine-tuning and joint reinforcement learning of dialogue policies and simulators was proposed, aimed at efficient domain adaptation.  More recently, a domain-independent transformer-based user simulator model was proposed by \cite{lin-etal-2021-domain}, demonstrating effective zero-shot transfer to unseen domains on the MultiWoz 2.1 corpus.  This was followed up by an approach that used the BART pre-trained language model to generate both user semantics and text \cite{Lin_etal_2022_sigdial}.

Adversarial learning has not been used before in the context of user simulation in dialogue, but some previous work has focused on using GANs in text generation \cite{demasson2019, betti-etal-2020-controlled, wu2021textgail} and dialogue generation \cite{li-etal-2017-adversarial, cui-etal-2019-dal, olabiyi-etal-2019-multi}.  In the context of task-oriented dialogue, adversarial learning has been used for reinforcement learning of dialogue policies, in which the discriminator estimates the rewards, based on human agent responses \cite{liu-lane-2018-adversarial}.

\section{Conclusion and future work}\label{sec:concl}

In this paper, we have presented a novel GAN-based architecture for data-driven user simulation in task-oriented dialogue systems.  The GAN consists of a Generator model for producing sequences of user dialogue acts, and a Discriminator model for classifying these sequences as simulated or real.  Various models have been trained and evaluated on a dialogue corpus, including two MLE baseline models and several GANs.  Furthermore, dialogue policies were trained and cross-evaluated with these simulator models, suggesting that out of the neural simulators, the best policies are obtained with an adversarially trained simulator, outperforming the best MLE model by 8.3\% success rate.  In a dialogue system user evaluation, policies from the best MLE, the best GAN, and a rule-based simulator were compared, showing that the GAN-based policy outperformed the MLE-based policy in terms of various subjective metrics.

Although the results on using GANs for semantic level user simulation are promising, we believe that models are likely to benefit even more from adversarial learning when generating both text and semantics.  The scope for generalisation beyond the training data is much larger when generating (longer) sequences of words rather than (shorter) sequences of user acts.  We are currently investigating how this approach can complement the use of pre-trained large language models.


\bibliographystyle{IEEEtran}
\bibliography{from_anthology,keizer-etal_usim_is2023}


%
%
%
%

\end{document}